\documentclass[letterpaper]{article} 
\usepackage{aaai2026}  
\usepackage{times}  
\usepackage{helvet}  
\usepackage{courier}  
\usepackage[hyphens]{url}  
\usepackage{graphicx} 
\usepackage{natbib}  
\usepackage{caption} 
\usepackage{algorithm}
\usepackage{algorithmic}

\usepackage{newfloat}
\usepackage{listings}
\DeclareCaptionStyle{ruled}{labelfont=normalfont,labelsep=colon,strut=off} 
\floatstyle{ruled}
\newfloat{listing}{tb}{lst}{}
\floatname{listing}{Listing}
\usepackage{booktabs}
\usepackage{amsmath}
\usepackage{amsfonts}
\usepackage{multirow}
\usepackage{comment}
\usepackage{cuted}
\usepackage{afterpage}

\title{3D-DRES: Detailed 3D Referring Expression Segmentation}
\author{
    Qi Chen\equalcontrib\textsuperscript{\rm 1},
    Changli Wu\equalcontrib\textsuperscript{\rm 1 2},
    Jiayi Ji\thanks{Corresponding author}\textsuperscript{\rm 1 3},
    Yiwei Ma\textsuperscript{\rm 1},
    Liujuan Cao\textsuperscript{\rm 1}
}
\affiliations{
    \textsuperscript{\rm 1}Key Laboratory of Multimedia Trusted Perception and Efficient Computing, Ministry of Education of China, Xiamen University, 361005, P.R. China

    \textsuperscript{\rm 2}Shanghai Innovation Institute
    
    \textsuperscript{\rm 3}National University of Singapore

    chenqi@stu.xmu.edu.cn, wuchangli@stu.xmu.edu.cn, jjyxmu@gmail.com, mayiwei@stu.xmu.edu.cn, caoliujuan@xmu.edu.cn
}

\usepackage{bibentry}

\begin{document}
\maketitle

\begin{figure*}[t]
\centering
\includegraphics[width=0.94\textwidth]{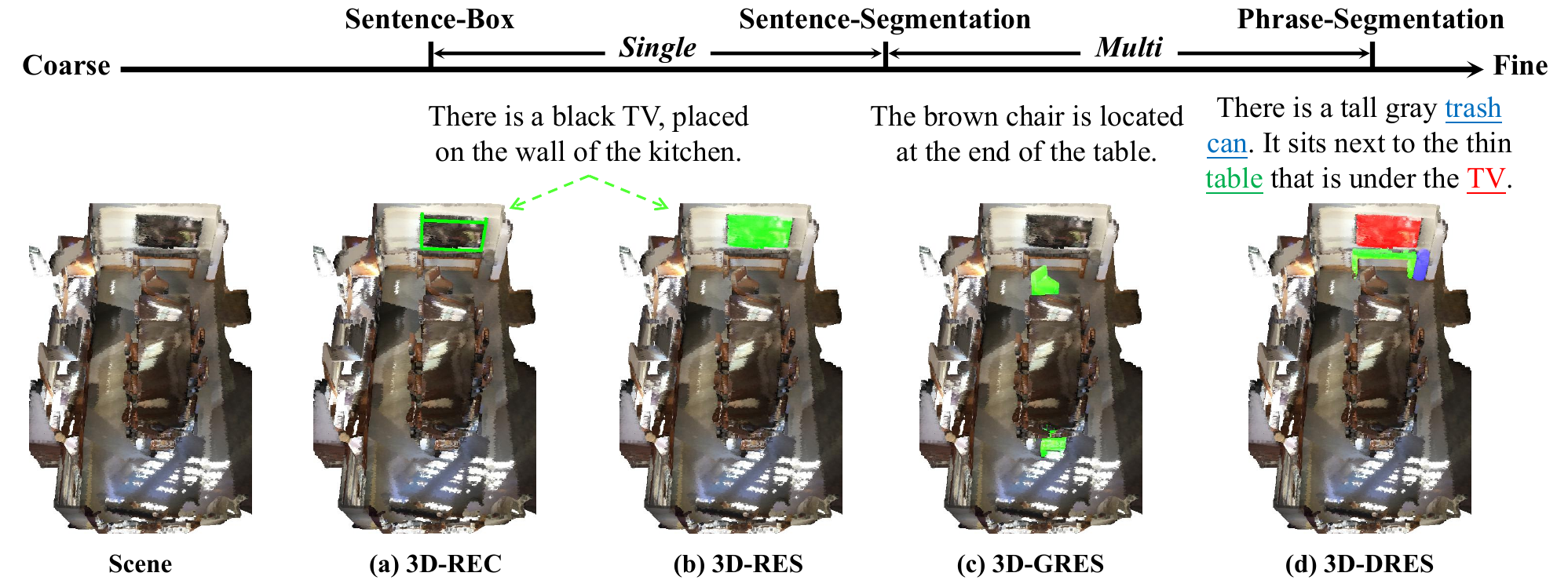} 
\caption{Illustration of 3D visual grounding tasks. (a) 3D Referring Expression Comprehension (3D-REC). (b) 3D Referring Expression Segmentation (3D-RES). (c) Generalized 3D Referring Expression Segmentation (3D-GRES). (d) Detailed 3D Referring Expression Segmentation (3D-DRES).}
\label{task_cmp}
\end{figure*}

\begin{abstract}
Current 3D visual grounding tasks only process sentence-level detection or segmentation, which critically fails to leverage the rich compositional contextual reasonings within natural language expressions. To address this challenge, we introduce Detailed 3D Referring Expression Segmentation (3D-DRES), a new task that provides a phrase to 3D instance mapping, aiming at enhancing fine-grained 3D vision-language understanding. To support 3D-DRES, we present DetailRefer, a new dataset comprising 54,432 descriptions spanning 11,054 distinct objects. Unlike previous datasets, DetailRefer implements a pioneering phrase-instance annotation paradigm where each referenced noun phrase is explicitly mapped to its corresponding 3D elements. Additionally, we introduce DetailBase, a purposefully streamlined yet effective baseline architecture that supports dual-mode segmentation at both sentence and phrase levels. Our experimental results demonstrate that models trained on DetailRefer not only excel at phrase-level segmentation but also show surprising improvements on traditional 3D-RES benchmarks. 

\end{abstract}

\begin{links}
     \link{Github}{https://github.com/80chen86/3D-DRES}
\end{links}

\section{Introduction}
\label{sec:intro}
Vision-language integration stands at the forefront of computer vision research, enabling machines to understand and reason about the visual world through natural language~\cite{vqa_1,vqa_2,res_2,img_cap_1,vl1,vl2}. In recent years, with the rapid advancement of 3D sensing technologies and deep learning models, 3D vision-language tasks have emerged as a prominent research focus in the community~\cite{3dcap_1,he2025refersplat,3dsurvey,vqa_4,3dqa_1, scanrefer, tgnn}. 
These tasks enable critical applications across robotics, autonomous navigation, mixed reality, and assistive technologies, where understanding the relationship between 3D environments and natural language is essential.

Among these, 3D visual grounding tasks are particularly crucial as they require localizing targets in 3D scenes based on textual instructions—a fundamental capability for embodied AI and autonomous systems. The field has evolved systematically, beginning with 3D-REC~\cite{scanrefer,3drec_7,3drec_8,3drec_9,3drec_10,3drec_11,3drec_12,3drec_16}, which localizes objects using coarse 3D bounding boxes and formulates the problem as coordinate regression (Fig.~\ref{task_cmp}-(a)). Subsequently, 3D-RES was introduced to address the need for finer-grained localization~\cite{tgnn,3dres_1,mcln,3dres_5}, requiring point-level segmentation and transforming the task into an expression-to-point matching problem (Fig.~\ref{task_cmp}-(b)). However, both tasks were constrained by their ability to handle only one-to-one mappings between sentences and objects, limiting their practical applicability in scenarios where instructions might refer to multiple objects or none at all. To address this limitation, 3D-GRES (Fig.~\ref{task_cmp}-(c)) expanded the task formulation to accommodate zero, one, or multiple targets per textual description~\cite{3dgres,ipdn, multi3drefer,3dgrec_1}. These developments have significantly advanced the field of 3D visual grounding, establishing it as a cornerstone of 3D scene understanding.

Despite these significant advancements, existing 3D visual grounding tasks still suffer from a prominent issue: the single-unit assumption (we define an ``unit" as one or more objects that do not need to be distinguished). Specifically, current tasks are limited to sentence-level segmentation or localization, meaning they focus only on a single unit described in a sentence, which greatly limits both practical applications and academic research. In real-world applications, completing user-issued commands often requires attention to all units mentioned. For example, in the common instruction, ``Put these clothes into the washing machine", both ``clothes" and ``washing machine" are essential units to consider. In academic research, the single-unit assumption hinders the comprehensive evaluation of models' fine-grained linguistic understanding and constrains the modeling of intra-sentence relationships and semantic structures. As illustrated in Fig.~\ref{task_cmp}-(d), traditional 3D-RES approaches provide no mechanism to determine whether a model correctly comprehends individual elements like ``table" and ``TV", even when localization appears successful for ``trash can". This limitation creates a critical gap in interpretability, as effective referring expression understanding inherently depends on contextual reasoning capabilities within the text. The need for a more granular, phrase-level understanding that would enable 3D vision-language models to develop robust contextual reasoning capabilities represents a significant unexplored opportunity in this domain.


Accordingly, we propose a novel task, Detailed 3D Referring Expression Segmentation (3D-DRES). This task requires models to focu all units within sentences. Specifically, as shown in Fig.\ref{task_cmp}-(d), the dataset provides the positions of all units to be segmented in the sentence (such as ``trash can", ``table", and ``TV"), and the model needs to generate the corresponding mask for each unit separately. The nature of the 3D-DRES task format inherently determines that models trained under this task focus more on the fine-grained semantic within sentences and is more aligned with practical applications. Meanwhile, the 3D-RES and the 3D-DRES can be mutually beneficial. We demonstrate this point with experiments in the subsequent sections.


A critical challenge in advancing this new paradigm is the absence of a suitable dataset. Creating such a resource is particularly challenging given the substantial costs associated with 3D annotation. To break through this dilemma, we invested approximately 600 hours to build DetailRefer, a new dataset built upon ScanRefer~\cite{scanrefer} through a combination of meticulous manual annotation and large language model~\cite{qwen} assistance. DetailRefer contains 54,432 descriptions covering 11,054 distinct objects, with an average text length of 24.9 tokens - significantly exceeding existing datasets (9.7-20.1 tokens). Unlike the sentence-segmentation annotation format prevalent in current datasets~\cite{scanrefer,referit3d,multi3drefer} where one sentence corresponds to one segmentation result, DetailRefer implements a pioneering phrase-segmentation format where each noun phrase corresponds to a distinct segmentation mask. This results in an unprecedented density of 2.9 masks per text. The dataset strategically incorporates 7.4\% ``Long" texts (exceeding 50 tokens) and numerous ``Complex" samples requiring segmentation of four or more noun phrases, establishing a robust framework for evaluating models' fine-grained linguistic understanding capabilities in 3D environments.

We observe that most existing models~\cite{tgnn,ipdn,mcln,unified} are designed for ``sentence-segmentation" sample formats and are unable to output multiple masks or specify masks for particular tokens. This limitation makes it impossible to apply current methods to the 3D-DRES task directly. Therefore, as initiators of this task, we propose DetailBase, a purposefully streamlined but effective baseline to establish the foundation for future research. Our design philosophy prioritizes simplicity to ensure high scalability and adaptability, while simultaneously demonstrating sufficient effectiveness to validate the task's potential. DetailBase features an elegant architecture that supports both sentence-level and phrase-level segmentation, with our comprehensive experiments confirming its efficacy across various evaluation dimensions. Importantly, our results demonstrate that training on this fine-grained task yields surprising improvements even on traditional 3D-RES benchmarks, suggesting that phrase-level understanding enhances overall spatial reasoning capabilities.

To sum up, our main contributions are as follows:

\begin{itemize}
    \item We introduce 3D-DRES, a novel fine-grained visual grounding task that is designed to enhance the ability to understand and localize textual context in 3D vision and language tasks.
    \item We have created a new dataset based on the Scannet indoor point cloud scenes, combining human annotations with large language models.
    \item We provide a simple, yet effective framework to serve as a foundational starting point for studying the 3D-DRES.
\end{itemize}

\section{Related Work}
\label{sec:related}

\subsection{3D Referring Expression Segmentation}

The 3D-RES task was first proposed by Huang et al.~\cite{tgnn} in 2021, aiming to segment the target object specified in the point cloud based on a given textual description. There are currently three main datasets for this task, namely ScanRefer~\cite{scanrefer}, Sr3D, and Nr3D~\cite{referit3d}. The text descriptions in these three datasets are all based on the indoor point cloud scene dataset Scannet~\cite{scannet}. In terms of details, the descriptions in ScanRefer and Nr3D are manually annotated by humans. Sr3D is a template based description generated based on the mutual positional relationship between objects, which has lower complexity.

In response to the shortcomings of 3D-RES tasks, namely the existence and uniqueness of target objects, Wu et al.~\cite{3dgres} proposed the 3D-GRES task, aimed at lifting the limitation on the number of target objects. The task currently only has one dataset, Multi3DRefer~\cite{multi3drefer}, proposed by Zhang et al. It modifies and enhances the text based on ScanRefer, so that the number of targets corresponding to the text is no longer fixed.

To address these two tasks, researchers have proposed numerous methods~\cite{3dres_1,3dres_3,unified,tgnn,mcln,3dgres} over time. However, most of these approaches are unfortunately limited to sentence-level segmentation, with their frameworks barely supporting fine-grained segmentation of noun phrases within sentences. Even when we extend our view to approaches~\cite{3drec_1,3drec_2,3drec_3,3drec_4,3drec_5,3drec_6,3drec_13,3drec_14,3drec_15,3drec_17} designed for the 3D-REC task, the situation does not improve much. Therefore, in this paper, we design a simple yet effective framework that supports both sentence-level and phrase-level segmentation. 

\subsection{2D Referring Expression Segmentation}

Compared to 3D-RES, 2D-RES~\cite{2dres_1, 2dres_2, 2dres_3, 2dres_4, 2dres_5} was proposed and studied by Hu et al.~\cite{first_res} as early as 2016, with the goal of segmenting target objects in images based on given text descriptions. The RES dataset in the 2D domain is more abundant compared to the 3D domain, with classic examples including RefCOCO, RefCOCO+~\cite{res_1}, and RefCOCOg~\cite{res_3}. Similarly, there is also GRES in the 2D field, with its dataset being gRefCOCO~\cite{gres}. In addition, Flickr30k Entities~\cite{flickr30k_entities, f30en_1, f30en_2, f30en_3, f30en_4, f30en_5} and Panoptic Narrative Grounding~\cite{png, png_1, png_2, png_3, png_4, png_5} are two datasets that are similar to this paper, both of which require segmentation of noun phrases in sentences. But the difference is that the former's supervisory signal is a box, while the latter is a text that corresponds to a panoramic description of an image. 

\section{DetailRefer Dataset}
\label{sec:dataset}

In this section, we first introduce the construction process of the DetailRefer (Sec.~\ref{dataset_creation}). Afterwards, introduce the relevant statistical data of the dataset and compare it with other datasets (Sec.~\ref{dataset_stat}). Finally, we introduce the 3D-DRES task definition and metrics based on this dataset (Sec.~\ref{task}).

\subsection{Dataset Creation}\label{dataset_creation}
Similar to Multi3DRefer~\cite{multi3drefer}, we modify and enhance the ScanRefer~\cite{scanrefer} dataset to derive DetailRefer. In the first phase, we first use a program to divide the descriptions in ScanRefer, grouping all descriptions of the same object together. Then, we ask the LLM to consolidate these sentences into a more comprehensive new description. 
It is worth noting that we do not directly use the texts from ScanRefer because each text typically involves fewer objects, which aligns less with our task objectives. After obtaining approximately 10,000 new descriptions, we manually annotate all noun phrases in each text to link them to specific objects in the 3D scene. During the annotation process, any inaccuracies in the descriptions are also corrected.

\begin{figure}[t]
\centering
\includegraphics[width=0.99\columnwidth]{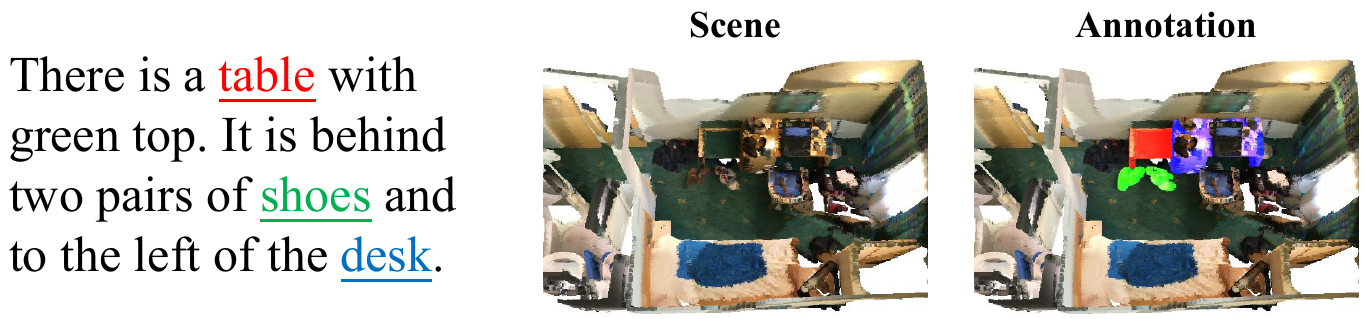} 
\caption{Example of DetailRefer.}
\label{example}
\end{figure}

\begin{figure}
\centering
\includegraphics[width=0.8\columnwidth]{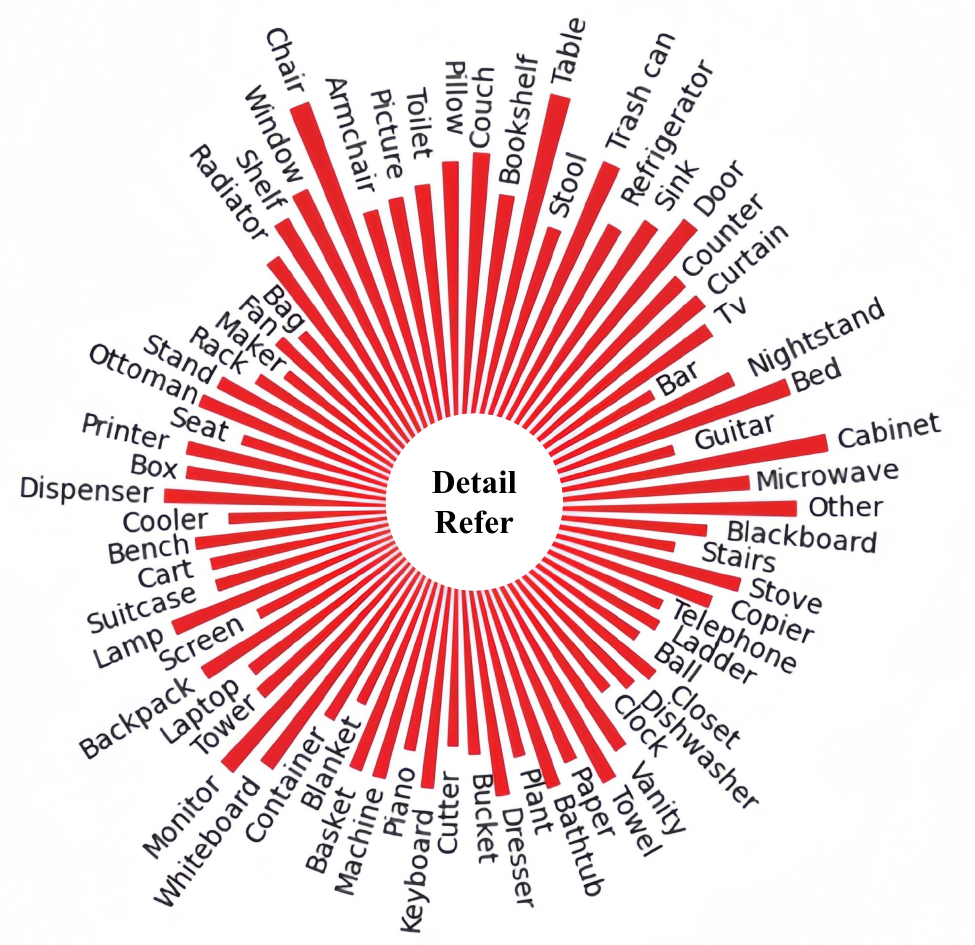} 
\caption{Category distribution of phrases in DetailRefer.}
\label{catagory_dis}
\end{figure}

After the first phase, we initially obtained a dataset that met our criteria, but it was relatively small in scale. Therefore, in the second phase, we needed to enhance and expand this dataset. Specifically, we transform the annotated text into a format that is easier for LLM to understand. This means appending a `()' immediately after each noun phrase, with the object IDs corresponding to that noun phrase enclosed within the `()'. We then fed these sentences into the LLM, instructing it to generate several different expressions while retaining the original semantic meaning and keeping the object IDs following the corresponding noun phrases. Using this method, we expanded the dataset to five times its original size, achieving a scale comparable to ScanRefer.

At this point, we have actually completed the initial construction of the dataset. However, considering that each text currently only describes a small area centered around a single object, we aim to obtain more challenging texts that cover larger areas. Thus, we traverse each object mentioned in the dataset, extract all texts from the first-phase dataset that involve that object, and input these texts into the LLM for integration to generate texts with broader descriptions. With this, the construction of the dataset is complete, and the full construction pipeline can be found in the appendix. We have included a sample from the dataset in the Fig.~\ref{example}.

\subsection{Dataset Statistics} \label{dataset_stat}

\begin{table}
    \centering
    \resizebox{0.85\linewidth}{!}{
        \begin{tabular}{c|cccc}
            \toprule
            Dataset & \textbf{Avg. length} & \textbf{Long} & \textbf{Avg. mask} & \textbf{Num}\\
            \midrule

            ScanRefer
            & 20.1 & 0.5\% & 1.0 & 51583 \\

            Sr3D
            & 9.7 & 0.0\% & 1.0 & 83572 \\

            Nr3D
            & 11.5 & 0.3\% & 1.0 & 41503 \\

            Multi3DRefer
            & 15.1 & 0.0\% & 1.0 & 61926 \\

            DetailRefer (ours)
            & 24.9 & 7.4\% & 2.9 & 54432 \\
            
            \bottomrule
        \end{tabular}
}
    \caption{Datasets comparison.}
    \label{tab:dataset_comparison}
\end{table}

DetailRefer contains a total of 54,432 descriptions, involving 11,054 different objects from the Scannet~\cite{scannet}. After dividing according to scenes in Scannet, the training, validation, and test sets contain 43,282, 5,398, and 5,752 descriptions, respectively. DetailRefer comprises 156636 noun phrases to be segmented. We present the category distribution of the objects corresponding to these phrases in Fig.~\ref{catagory_dis}. Additionally, among all the noun phrases to be segmented, 15,001 phrases involve multiple objects; in all texts, long texts (more than 50 tokens) account for 7.4\%.

To facilitate comparison, we have compiled the various statistics of mainstream datasets~\cite{scanrefer, referit3d, multi3drefer} in Tab~\ref{tab:dataset_comparison}. It can be seen that compared to other datasets, our dataset excels in average text length. Additionally, it is observed that long texts exceeding 50 tokens are extremely rare, almost non-existent, in other datasets. However, in our dataset, 7.4\% of the texts are long, which better tests the model's understanding of natural language. Furthermore, since other datasets correspond to sentence-level segmentation tasks, each text only has one mask. In contrast, our dataset corresponds to noun phrase-level segmentation tasks, thus each text averages 2.9 masks. From the comparison of text quantities, it is evident that our dataset is comparable in scale to existing mainstream datasets.

\subsection{3D-DRES Task} \label{task}

\subsubsection{Task Definition}The Detailed 3D Referring Expression Segmentation (3D-DRES) task aims to segment out masks corresponding to each target phrase given in a sentence from a point cloud scene. Specifically, first, given a point cloud scene  $P\in\mathbb{R}^{N_{p}\times f}$, where $N_{p}$ represents the number of points and $f$ is the feature length (including coordinates XYZ, color RGB, etc.). Secondly, given a textual description $T\in\mathbb{R}^{L}$, where $L$ is the number of tokens in the text. Finally, given a set of indices $I=\{i_{1},i_{2},\ldots,i_{k}\}$, corresponding to the positions of $k$ nouns in the text that need segmentation, the model is required to output point cloud scene masks $Mask\in\mathbb{R}^{k\times N_{p}}$ for all nouns.

\subsubsection{Metrics}First, we focus on phrase-level mean IoU (mIoU), Acc@0.25, and Acc@0.5. At the phrase level, mIoU is the average of the IoUs calculated for all terms to be segmented, while Acc@0.25 and Acc@0.5 represent the proportion of segmented terms with an IoU greater than 0.25 and 0.5 out of all segmented terms. Their formulas are as follows:
\begin{equation} \label{miou}
    \text{mIoU} = \frac{\sum_{i=1}^{M}\sum_{j=1}^{K_{i}}IoU_{i,j}}
    {\sum_{i=1}^{M}K_{i}},
\end{equation}
\begin{equation}
    \text{Acc@}t = \frac{\sum_{i=1}^{M}\sum_{j=1}^{K_{i}}B_{t}(IoU_{i,j})}
    {\sum_{i=1}^{M}K_{i}}, t\in\{0.25,0.5\}
\end{equation}
where $M$ represents the number of descriptions in the dataset, $K_{i}$ denotes the number of nouns that need to be segmented in the $i$-th description, $IoU_{i,j}$ represents the IoU value corresponding to the $j$-th noun in the $i$-th description, and $B_{t}(\cdot)$ indicates that it returns 1 if the parameter is greater than $t$, otherwise it returns 0. Additionally, we also focus on the sentence-level mean IoU (mIoU-S), whose calculation formula is as follows:
\begin{equation}
    \text{mIoU-S} = \frac{1}{M}\sum_{i=1}^{M}\frac{1}{K_{i}}\sum_{j=1}^{K_{i}}IoU_{i,j},
\end{equation}
where the meanings of all symbols are the same as in Eq.~\ref{miou}. The mIoU-S reflect the model's understanding capability at the sentence level. 

\begin{figure*}[t]
\centering
\includegraphics[width=0.8\textwidth]{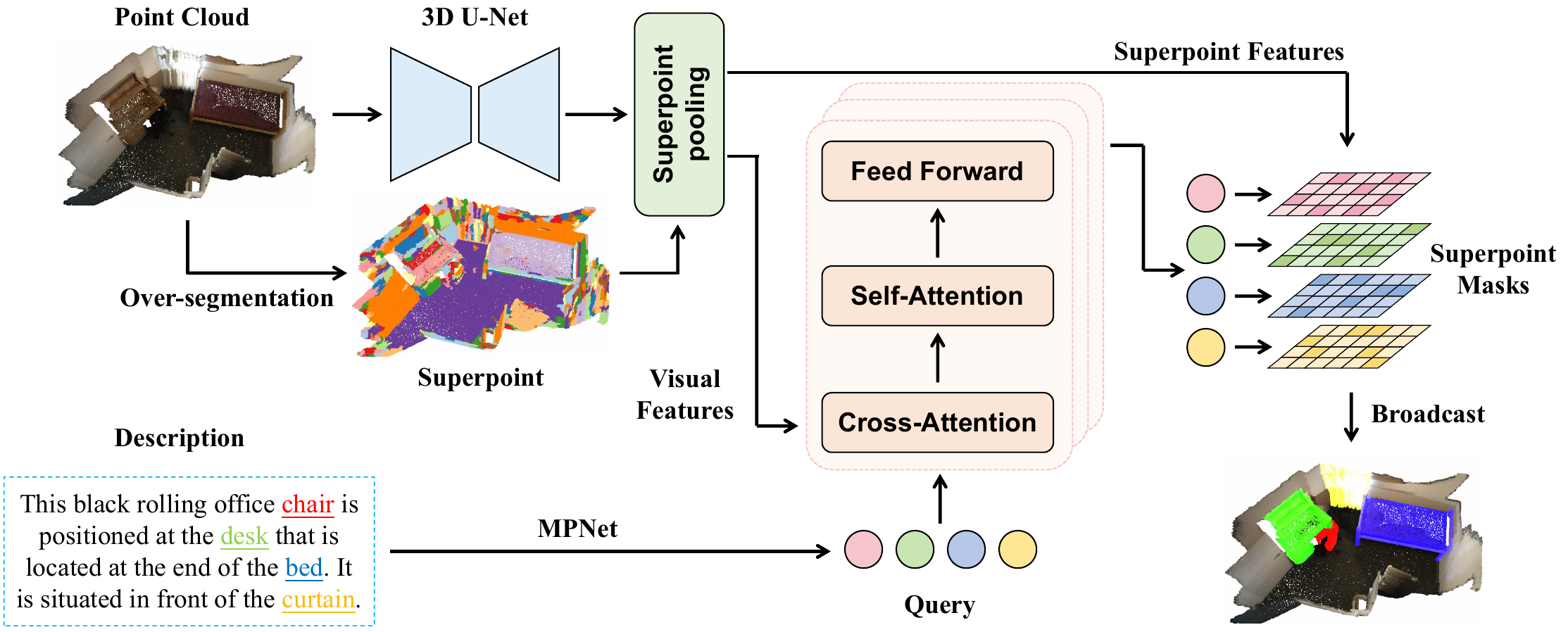} 
\caption{The overview of the Detailed 3D Referring Expression Segmentation Baseline (DetailBase). }
\label{detailbase}
\end{figure*}

Finally, we also pay special attention to the model's performance on long texts and complex scene descriptions. Specifically, we define texts with more than 50 tokens as long texts, and evaluate the metrics of the model on such texts to reflect its understanding capability of long texts. Meanwhile, we define texts that require segmentation of four or more phrases as complex scene descriptions, and test the model's metrics on these texts to reflect its understanding capability of complex scene descriptions.

\section{Baseline of 3D-DRES}

Existing 3D-RES methods are based on the assumption of ``sentence-level segmentation," which typically cannot output multiple masks or specify masks for particular tokens. This limitation means that they cannot be directly applied to our new task. Therefore, in this section, as the proposers of 3D-DRES, we introduce a simple, effective, and highly extensible framework to serve as a foundational starting point for investigating this task. We name this framework DetailBase, and an overview is shown in Fig.~\ref{detailbase}.

According to the task definition in Sec.~\ref{task}, the inputs for this task include a point cloud scene $P\in\mathbb{R}^{N_{p}\times f}$, a textual description $T$, and the position $I$ of the noun to be segmented. We first obtain point-level features $F_{p}$ by feeding the point cloud (here, we only use coordinates XYZ and color RGB as the initial features for each point) into a 3D U-Net network~\cite{3dunet}. Given that the number of these features is excessively large, we adopt the same approach as previous models~\cite{3dstmn,spformer}, namely superpoint pooling, to simplify them. Specifically, we perform an unsupervised oversegmentation on the $P$, resulting in $N_{s}$ superpoints $\{SP_{i}\}_{i=1}^{N_{s}}$, where $N_{p}>>N_{s}$~\cite{sp}. Afterward, we average the features of all points belonging to the same superpoint. Finally, through two independently weighted linear transformations, the pooling features are converted into visual features for multimodal information fusion and superpoint features for predicting the mask. The entire process is formulated as follows:
\begin{equation}
    F_{pool} = \text{SPPool}(F_{p}, P),
\end{equation}
\begin{equation}
    F_{v} = F_{pool}W_{1}, F_{sp} = F_{pool}W_{2},
\end{equation}
where $F_{pool}\in\mathbb{R}^{N_{s}\times c}$ denotes the pooling features, $\text{SPPool}(\cdot)$ represents the superpoint pooling operation, $F_{v}\in\mathbb{R}^{N_{s}\times d}$ indicates the visual features, $F_{sp}\in\mathbb{R}^{N_{s}\times d}$ denotes the superpoint features, and $W_{1},W_{2}\in\mathbb{R}^{c\times d}$ are both randomly initialized learnable parameters.

For a given text $T$, special tokens are added to both the beginning and the end of the text before it is input into the MPNet network~\cite{mpnet} to obtain token features. Given that this task requires the model to be capable of segmenting tokens at specified positions, we use the token features to generate the initial query $Q_{0}$:
\begin{equation}
    Q_{0} = EW_{3}
\end{equation}

\begin{table*}[t]
    \centering
    \resizebox{0.9\textwidth}{!}{
        \begin{tabular}{c|cccc|cccc|cccc}
            \toprule
            & \multicolumn{4}{c|}{Long} 
            & \multicolumn{4}{c|}{Complex} 
            & \multicolumn{4}{c}{Overall} \\
            \multirow{-2}{*}{Method} 
            & 0.25 & 0.5 & mIoU-S & mIoU
            & 0.25 & 0.5 & mIoU-S & mIoU
            & 0.25 & 0.5 & mIoU-S & mIoU \\
            \hline

            \multicolumn{13}{c}{{Val}}\\
            \hline

            PNG ~\cite{png}
            & 50.3 & 34.0 & 35.8 & 35.1
            & 52.7 & 35.4 & 37.0 & 36.4
            & 57.2 & 41.9 & 42.6 & 41.3 \\

            3D-STMN ~\cite{3dstmn}
            & 63.8 & 46.0 & 46.4 & 45.2
            & 65.4 & 48.8 & 48.0 & 47.1
            & 71.8 & 55.8 & 53.8 & 52.7 \\


            DetailBase (Ours) 
            & \textbf{67.3} & \textbf{49.0} & \textbf{50.2} & \textbf{48.9}
            & \textbf{70.3} & \textbf{52.0} & \textbf{52.5} & \textbf{51.3}
            & \textbf{73.9} & \textbf{58.4} & \textbf{56.3} & \textbf{55.4} \\
            \hline

            \multicolumn{13}{c}{{Test}}\\
            \hline

            PNG ~\cite{png}
            & 52.0 & 35.6 & 37.9 & 36.5
            & 53.8 & 37.3 & 38.5 & 37.7
            & 56.3 & 40.7 & 41.1 & 40.4 \\

            3D-STMN ~\cite{3dstmn}
            & 67.1 & 48.9 & 48.7 & 47.6
            & 69.9 & 51.8 & 51.1 & 50.0
            & 71.9 & 54.8 & 53.1 & 52.5 \\

            DetailBase (Ours) 
            & \textbf{71.1} & \textbf{52.8} & \textbf{52.5} & \textbf{51.5}
            & \textbf{73.5} & \textbf{55.8} & \textbf{54.8} & \textbf{53.8}
            & \textbf{74.8} & \textbf{58.5} & \textbf{56.2} & \textbf{55.7} \\
            \bottomrule
        \end{tabular}
    }

    \caption{3D-DRES results on DetailRefer dataset, showing mIoU, mIoU-S and accuracy at IoU thresholds of 0.25 and 0.5. ``Long" indicates samples exceeding 50 tokens, while ``Complex" refers to samples with four or more noun phrases to segment.}
    \label{tab:detailrefer_benchmark}
\end{table*}

where $E\in\mathbb{R}^{(L+2)\times e}$ represents the token features, and $W_{3}\in\mathbb{R}^{e\times d}$ denotes the learnable parameters. Subsequently, $Q_{0}$ is fed into the decoder, where it first uses cross-attention~\cite{attention} to integrate information from the visual modality, then employs self-attention to focus on the internal information within the sentence, and finally passes through a feed-forward network for nonlinear transformation. Additionally, we adopt a multi-layer architecture in series, meaning there are multiple such Cross-Self-FFN structures. For generality, the formula here is given for the $i$-th layer as an example:
\begin{equation}
    Q_{i} = \text{FFN}(\text{Self}(\text{Cross}(Q_{i-1},F_{v}))), i\in\{1,\ldots,N_{l}\},
\end{equation}
where $Q_{i-1}$ represents the query output from the previous layer and $N_{l}$ denotes the number of model layers. Finally, we compute the affinity between the query output from the last layer and the superpoint features, and then binarize this affinity to obtain the superpoint mask corresponding to the query. The superpoint mask can be broadcast to obtain point-level masks. For the sentence-level segmentation, use the mask corresponding to the [CLS] token as the result.

During the model training phase, we compute the superpoint mask for the query output from each layer and then calculate the loss against the ground truth. Here, we use the classic BCE loss and Dice loss~\cite{dice}. Additionally, similar to previous works~\cite{3dstmn,ipdn}, we add an auxiliary Score loss. The final loss is formulated as follows:
\begin{equation}
    L_{total} = \sum_{i=0}^{N_{l}}\lambda_{1}L_{BCE}^{i} + \lambda_{2}L_{Dice}^{i}+\lambda_{3}L_{Score}^{i}
\end{equation}
where the superscript $i$ denotes the layer number, and $\lambda_{1}$, $\lambda_{2}$, $\lambda_{3}$ are hyperparameters. $L_{total}$, $L_{BCE}$, $L_{Dice}$, $L_{Score}$ represent the total loss, BCE loss, Dice loss, and Score loss, respectively. During the model inference phase, only the output from the last layer is used as the final result.

\section{Experiments}

\subsection{Experiment Settings}
We utilize a Sparse 3D U-Net~\cite{3dunet} for extracting features from point clouds and a pre-trained MPNet~\cite{mpnet} for text feature extraction. The initial learning rate is set to 0.0001, which we decay at epoch {26, 34, 42} with a decay rate of 0.5. The number of model layers $N_{l}$ is set to 6. In the hyperparameters, $\lambda_{1}$, $\lambda_{2}$, and $\lambda_{3}$ are set to 1, 1, and 0.5, respectively. The batch size is set to 16. All experiments are trained on an NVIDIA GeForce RTX 3090 GPU.

\subsection{Quantitative Results}
Given that there currently exists no model that can directly adapt to the 3D-DRES task, we made appropriate adjustments to two existing models, namely PNG~\cite{png} and 3D-STMN~\cite{3dstmn}, to fit our task. The PNG model originates from the 2D domain's Panoptic Narrative Grounding task, which is quite similar to our task. We first changed its input by using point clouds instead of images, then utilized SPFormer~\cite{spformer} to extract instance proposals and averaged the point features within each instance as the instance feature. We also modified its result matching method, changing from selecting the candidate with the highest matching similarity to candidates with matching similarity above a certain threshold. This adjustment was made due to cases in our task where one noun phrase corresponds to multiple instances. For 3D-STMN, we altered its supervision method and result generation approach to align with DetailBase. Additionally, in the original 3D-STMN, the 3D-UNet was frozen. To ensure fairness, we have opted to include it in the training process here.

In Tab.~\ref{tab:detailrefer_benchmark}, we present the performance of each model on both the validation set and the test set. It can be observed that although the modified PNG model achieved an miou of 40.4 on the test set, it still significantly lags behind other models. The reason for this situation is that the current effectiveness of 3D instance segmentation networks is not ideal; such two-stage models generally perform weaker than one-stage models. The modified 3D-STMN demonstrated its capabilities in this task, but due to the low compatibility of its designed modules with the task, its performance fell behind our DetailBase. DetailBase achieved an miou of 55.7 on the test set, a result that lays a suitable foundation for the further development of 3D-DRES methods. Moreover, our model framework is relatively simple and highly scalable, making it very appropriate as an initial approach for this task.

Phrase-level segmentation emphasizes fine-grained semantic understanding, whereas sentence-level segmentation focuses more on holistic comprehension. These two are not mutually exclusive; instead, they are mutually beneficial. We conducted joint training experiments to substantiate this viewpoint. By treating the [CLS] token (the root node in 3D-STMN) as the ``noun phrase" to be segmented for 3D-RES tasks, we unified the formats of both tasks. We performed separate training on ScanRefer~\cite{scanrefer} and DetailRefer datasets, as well as joint training across both datasets, reporting the mIoU of the models on the validation set in the Tab.~\ref{multi_task}. It is evident that joint training yields superior results compared to separate training for both 3D-STMN and DetailBase. Notably, joint training significantly enhances performance on 3D-RES, improving scores by 2.8 points on DetailBase and up to 3.2 points on 3D-STMN. In conclusion, our task not only holds its unique value but also complements traditional tasks synergistically.

\begin{table}
    \centering
    \resizebox{0.99\columnwidth}{!}{
        \begin{tabular}{c|cc|cc}
            \toprule  
            & \multicolumn{2}{c|}{\textbf{DetailBase}} 
            & \multicolumn{2}{c}{\textbf{3D-STMN}\cite{3dstmn}}\\
            \multirow{-2}{*}{\textbf{Dataset}}
            & 3D-RES & 3D-DRES  
            & 3D-RES & 3D-DRES \\
            \midrule

            Scanrefer & 44.0 & - & 41.8 & - \\

            DetailRefer & - & 55.4 & - & 52.7 \\

            Both & 46.8 & 56.8 & 45.0 & 53.3 \\
            
            \bottomrule
        \end{tabular}
    }

    \caption{Results of separate training and joint training. 
    }
    \label{multi_task}
\end{table}

\begin{table}
    \centering
    \resizebox{0.8\columnwidth}{!}{
        \begin{tabular}{c|cc|ccc}
            \toprule  & &
            & \multicolumn{3}{c}{\textbf{Overall}} \\
            \multirow{-2}{*}{$N_{l}$}
            & \multirow{-2}{*}{\textbf{Long}} & \multirow{-2}{*}{\textbf{Complex}}  
            & 0.25 & 0.5 & mIoU \\
            \midrule

            1 & 46.7 & 49.0 & 74.4 & 54.1 & 53.1 \\

            3 & 48.6 & 50.8 & 73.8 & 58.2 & 55.1 \\

            6 & \textbf{48.9} & 51.3 & \textbf{73.9} & 58.4 & 55.4 \\

            9 & 48.2 & \textbf{51.4} & 71.8 & \textbf{59.7} & \textbf{55.5} \\
            
            \bottomrule
        \end{tabular}
    }

    \caption{Ablation study on number of model layers.}
    \label{tab:layer_num}
\end{table}

\begin{table}
    \centering
    \resizebox{0.9\columnwidth}{!}{
        \begin{tabular}{cc|ccc}
            \toprule
            \textbf{Multi\_layer} & \textbf{Score} 
            & \textbf{Long} & \textbf{Complex} & \textbf{Overall}\\
            \midrule

            $\times$ & $\times$ & 45.0 & 46.3 & 50.5 \\

            $\times$ & \checkmark & 44.5 & 46.5 & 50.8 \\

            \checkmark & $\times$ & 48.4 & 50.9 & 55.0 \\

            \checkmark & \checkmark & \textbf{48.9} & \textbf{51.3} & \textbf{55.4} \\
            
            \bottomrule
        \end{tabular}
    }

    \caption{Ablation on loss. `Multi\_layer' indicates that each layer is supervised. `Score' refers to auxiliary score loss.}
    \label{tab:loss}
\end{table}

\begin{figure}[t]
\centering
\includegraphics[width=0.99\columnwidth]{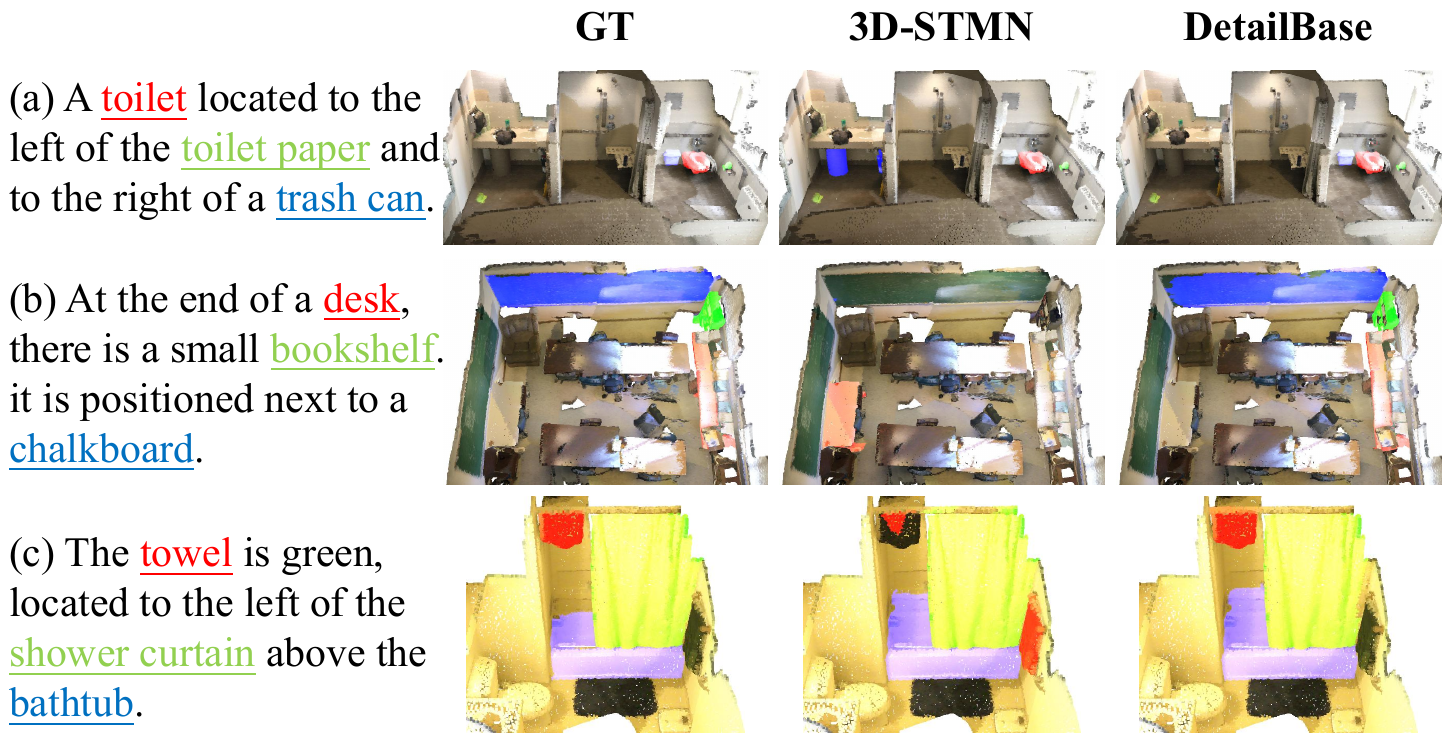} 
\caption{Comparison of visualization results. The 3D-STMN results are the model predictions after adaptation. }
\label{qualitative_result}
\end{figure}

\subsection{Ablation Study}
Ablation experiments are all conducted on the validation set.

We conducted an ablation study on the parameter $N_{l}$, which indicates the number of layers in the model. We present the mIoU for the `long' and `complex' subset, as well as the Acc@0.25, Acc@0.5, and mIoU on the entire validation set in Tab.~\ref{tab:layer_num}. It can be observed that with only one layer, due to having fewer parameters and insufficient fitting capability, the model performed poorly, achieving an mIoU of only 53.1. When the number of layers reached three, there was a noticeable improvement in performance, with the mIoU increasing to 55.1. Beyond this point, the performance gains from increasing the number of layers diminished. 
Taking into account both model performance and complexity, six layers is the optimal choice.

We conducted an ablation study on the model's loss and presented the mIoU in Tab.~\ref{tab:loss}. In this context, ``Multi\_layer" indicates that each layer is supervised; if not used, supervision only occurs at the final layer. ``Score" refers to Score loss, which employs an MLP to predict the IoU. As shown in the table, applying supervision at every layer has a significant impact, improving the mIoU on the overall validation set by nearly five points. Regarding the Score loss, although its enhancement is relatively minor, as an auxiliary los, its computational cost is negligible. Therefore, adding the Score loss is a beneficial and harmless operation. 

\begin{figure}[t]
\centering
\includegraphics[width=0.99\columnwidth]{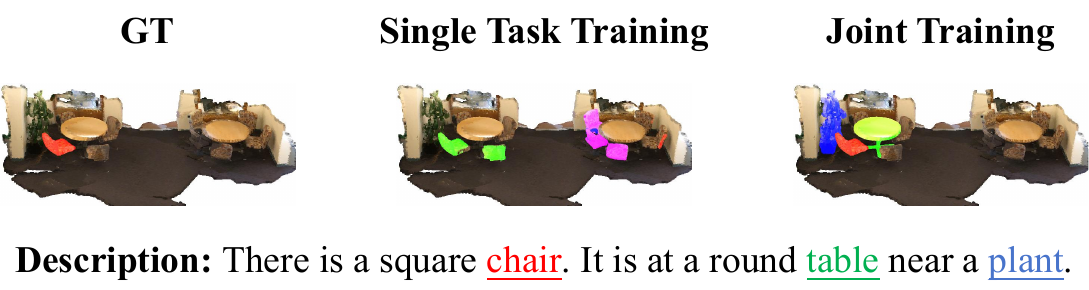} 
\caption{Visual performance of 3D-STMN on the ScanRefer dataset under different training methods.}
\label{fig:task_compare}
\end{figure}

\subsection{Qualitative Result}
We present some visualization results in Fig.~\ref{qualitative_result} to enable a more intuitive perception of the advantages of the 3D-DRES. As shown in (a), if it were a traditional 3D-RES, the text would refer to the object ``toilet'' and both models successfully segmented the target. 
It is difficult to assess a model's comprehension of the entire text at a fine-grained level, as the results are only evaluated based on a single entity within the text. However, under our 3D-DRES setting, we can observe the model's overall understanding of the text in a more detailed manner. For example, 3D-STMN demonstrates a completely incorrect understanding of text (b), whereas its comprehension of text (c) is partially correct.
Through detailed analysis of model capabilities, we can better define the optimization directions for the model.

Additionally, we visualized the 3D-RES capabilities of 3D-STMN~\cite{3dstmn} under different training methods, with one result shown in the Fig.~\ref{fig:task_compare} (more results in the appendix). It can be observed that when trained solely on the 3D-RES task, the model's fine-grained text understanding is poor. However, under joint training, the model's ability to capture fine-grained information within sentences significantly improves, enabling it to accurately locate auxiliary objects and thereby pinpoint the correct target.

\section{Analysis of 3D-DRES}
In this section, we briefly analyze the 3D-DRES task and summarize its challenges. A more detailed analysis, along with visual examples, is provided in Sec. 3 of the Appendix.


In our task, since the nouns to be segmented are specified and the variety of 3D scene objects is relatively small, the set of source noun phrases for generating segmentation kernels is not large. Additionally, approximately 10\% of the noun phrases in our dataset correspond to multiple target objects. These two factors contribute to the model's difficulty in distinguishing between instances, especially when two objects of the same category are in close proximity. Distinguishing instances and determining the exact number of instances to be segmented present a significant challenge in this task.



Regarding the DetailBase framework, there is a category of text that poses significant challenges: texts involving instance-level clues. This is because the single-stage framework operates only at the superpoint level and lacks the ability to perceive information at the instance level. Additionally, the number of superpoints themselves limits their interactions, causing each superpoint to act almost as an independent entity. How to effectively utilize instance-level cues is also a major challenge in this task. 

Finally, long texts are a distinctive feature of our dataset and also pose a challenge. Long texts imply more complex sentence structures, requiring the model to have a stronger ability to understand context.

\section{Conclusion}
In this paper, we introduce a novel task, 3D-DRES, which requires models to segment all noun phrases mentioned in sentences into corresponding masks. To support this task, we have constructed a new dataset, DetailRefer, featuring fine-grained annotations, combining both human effort and LLM. As the proposers of this task, we provide a highly scalable framework as a baseline for future researchers.

\section*{Acknowledgements}

This work was supported by the National Science Fund for Distinguished Young Scholars (No.62025603, No.62525605), the National Natural Science Foundation of China (No. U21B2037 , No. 62302411) and China Postdoctoral Science Foundation (No. 2023M732948).

\bibliography{aaai2026}

\end{document}